\documentclass[letterpaper,11pt]{article}
\usepackage[margin=1in]{geometry}

\usepackage[utf8]{inputenc} 
\usepackage[T1]{fontenc}    
\usepackage{hyperref}       
\usepackage{url}            
\usepackage{booktabs}       
\usepackage{amsfonts}       
\usepackage{nicefrac}       
\usepackage{microtype}      
\usepackage{amsmath, amsthm}
\usepackage{xcolor}
\usepackage[inline]{enumitem}
\usepackage{appendix}
\usepackage{natbib}
\newcommand{\D}{\mathcal{D}}
\newcommand{\A}{\mathcal{A}}
\newcommand{\B}{\mathcal{B}}
\newcommand{\LL}{\mathcal{L}}
\newcommand{\pbm}{\mathbb{P}}
\newcommand{\E}{\mathbb{E}}
\newcommand{\W}{\mathcal{W}}
\newcommand{\disc}{D_{\omega}}
\newcommand{\gene}{G_{\theta}}
\newcommand{\gt}{g_{\theta}}
\newcommand{\go}{g_{\omega}}
\newcommand{\gtij}{g_{\theta}^{i,j}}
\newcommand{\goij}{g_{\omega}^{i,j}}
\newcommand{\gtr}{g_{\theta}^{I,J}}
\newcommand{\gor}{g_{\omega}^{I,J}}
\newcommand{\gtb}{g_{\theta}^{\mathcal{B}}}
\newcommand{\gob}{g_{\omega}^{\mathcal{B}}}
\newcommand{\gtbk}{g_{\theta}^{I_k,J_k}}
\newcommand{\gobk}{g_{\omega}^{I_k,J_k}}
\newcommand{\sigt}{\Sigma_\theta}
\newcommand{\sigo}{\Sigma_\omega}
\newcommand{\params}{\theta,\omega}

\newcommand{\R}{\mathbb{R}}
\newcommand{\X}{\mathcal X}

\newcommand{\lrt}{\eta^\theta}
\newcommand{\lro}{\eta^\omega}

\newcommand{\ICal}{\mathcal I}
\newcommand{\II}{\mathbb I}
\newcommand{\N}{\mathbb N}
\newcommand{\CCal}{\mathcal C}

\newtheorem{remark}{Remark}[section]

\newtheorem{theorem}{Theorem}[section]

\newtheorem{lemma}{Lemma}[section]

\newtheorem{definition}{Definition}[section]

\title{SDE approximations of GANs training and its long-run behavior}

\author{Haoyang Cao\footnote{\'Ecole Polytechnique, Route de Saclay, 91128, Palaiseau Cedex, France, haoyang.cao@polytechnique.edu}\and
 Xin Guo\footnote{Department of Industrial Engineering and Operations Research, University of California, Berkeley, Berkeley, CA 94720, email: xinguo@berkeley.edu}}

\date{\today}

\begin{document}

\maketitle

\begin{abstract}
This paper analyzes the training process of GANs via stochastic differential equations (SDEs). It first establishes SDE approximations for the training of GANs under stochastic gradient algorithms, with precise error bound analysis. It then describes the long-run behavior of GANs training via the invariant measures of its SDE approximations under proper conditions. This work builds theoretical foundation for GANs training and provides analytical tools to study its evolution and stability. 
\end{abstract}

\section{Introduction}

Generative adversarial networks (GANs) introduced in \citet{Goodfellow2014}  are generative models between two competing neural networks: a generator network $G$ and a discriminator network $D$. The generator network $G$ attempts to fool the discriminator network by converting random noise into sample data, while the discriminator network $D$ tries to identify whether the input sample is fake or true. 

After being introduced to the machine learning community, the popularity of GANs has grown exponentially with a wide range of applications, including high resolution image generation \citep{denton2015deep,radford2015unsupervised}, image inpainting \citep{yeh2016semantic}, image super-resolution \citep{ledig2016others}, visual manipulation \citep{zhu2016generative}, text-to-image synthesis \citep{reed2016generative}, video generation \citep{vondrick2016generating}, semantic segmentation \citep{luc2016semantic}, and abstract reasoning diagram generation \citep{ghosh2016contextual}; in recent years, GANs have attracted substantial amount of attention in financial industry for financial time series generation \citep{Wiese2019, Wiese2020, Takahashi2019, Zhang2019}, asset pricing \citep{Chen2019}, market simulation \citep{coletta2021towards, storchan2021learning} and so on. Despite the empirical success of GANs, there are well-recognized issues in GANs training, such as the vanishing gradient when the discriminator significantly outperforms the generator \citep{Arjovsky2017a}, the mode collapse where the generator cannot recover a multi-model distribution but only a subset of the modes and this issue is believed to be linked with gradient exploding \citep{Salimans2016}. 

In response to these 
issues, there has been a growing research interest in the theoretical understanding of GANs training. In \citet{Berard2020} the authors proposed a novel visualization method for the GANs training process through the gradient vector field of loss functions.
In a deterministic GANs training framework, \citet{Mescheder2018} demonstrated that regularization improved  the convergence performance of GANs. \citet{Conforti2020} and \citet{Domingo-Enrich2020} analyzed a generic zero-sum minimax game including that of GANs,
and connected the mixed Nash equilibrium  of the game with the invariant measure of Langevin dynamics. In addition, various approaches have been proposed for amelioration of the aforementioned issues in GANs training, including  different choices of network architectures, loss functions, and regularization. See for instance, a comprehensive survey on these techniques \citep{Wiatrak2019} and the references therein.

\paragraph{Our work.} 
This paper focuses on analyzing the training process of GANs via a stochastic differential equations (SDEs) approach. It first establishes SDE approximations for the training of GANs under SGAs, with precise error bound analysis. It then describes the long-run behavior of GANs training via the invariant measures of its SDE approximations under proper conditions. This work builds theoretical foundation for GANs training and provides analytical tools to study its evolution and stability. In particular,
\begin{itemize}
\item[a)] the SDE approximations characterize precisely the distinction between GANs with alternating update and GANs with simultaneous update, in terms of the interaction between the generator and the discriminator; 
\item[b)] the drift terms in the SDEs show the direction of the parameters evolution; the diffusion terms prescribes the ratio between the batch size and the learning rate in order to modulate the fluctuations of SGAs in GANs training; 
\item[c)] regularity conditions for the coefficients of the SDEs provide constraints on the growth of the loss function with respect to the model parameters, necessary for avoiding the explosive gradient encountered in the training of GANs; they also explain mathematically some well known heuristics in GANs training, and confirm the importance of appropriate choices for network depth and of the introduction of gradient clipping and gradient penalty;
\item[d)] the dissipative property of the training dynamics in the form of SDE ensures the existence of the invariant measures, hence the steady states of GANs training in the long-run; it underpins the practical tactic of adding regularization term to the GANs objective to improve the stability of training;
\item[e)] further analysis of the invariant measure for the coupled SDEs gives rise to a fluctuation-dissipation relations (FDRs) for GANs. These FDRs reveal the trade-off of the loss landscape between the generator and the discriminator and can be used to schedule the learning rate.
\end{itemize}
\paragraph{Related works.}
Our analysis on the approximation and the long-run behavior of  GANs training is inspired by  \citet{Li2019} and \citet{Liu2019b}. The former established the SDE approximation for the parameter evolution in SGAs applied to pure minimization problems (see also \citet{Hu2019a} on the similar topic); the latter surveyed theoretical analysis of deep learning from two perspectives: propagation of chaos through neural networks and training process of deep learning algorithms. 
Among other related works on the theoretical understanding of GANs, in \citet{Genevay2017}, the authors reviewed the connection between GANs and the dual formulation of optimal transport problems; \citet{Luise2020} studied the interplay between the latent distribution and generated distribution in GANs with optimal transport-based loss functions;  \citet{Conforti2020} and \citet{Domingo-Enrich2020} focused on the equilibrium of the
minimax game and its connection with Langevin dynamics; \citet{Cao2020} studied the connection between GANs and mean-field games. Our focus is the GANs training process: we establish precise error bounds for the SDE approximations, study the long-run behavior of GANs training via the invariant measures of the SDE approximations, and analyze their implications for resolving various challenges in GANs.
 
\paragraph{Notations.}
Throughout this paper, the following notations will be adopted. 
\begin{itemize}
\item $\R^d$ denotes a $d$-dimensional Euclidean space, where $d$ may vary from time to time.
\item The transpose of a vector $x\in\R^d$ is denoted by $x^T$ and the transpose of a matrix $A\in\R^{d_1\times d_2}$ is denoted by $A^T$. 
\item Let $\mathcal X$ be an arbitrary nonempty subset of $\R^d$, the set of $k$ continuously differentiable functions over some domain $X$ is denoted by $\mathcal C^k(\mathcal X)$ for any nonnegative integer $k$. In particular when $k=0$, $\mathcal C^0(\mathcal X)=\mathcal C(\mathcal X)$ denotes the set of continuous functions. 
\item Let $J=(J_1,\dots,J_d)$ be a $d$-tuple multi-index of order $|J|=\sum_{i=1}^dJ_i$, where $J_i\geq0$ for all $i=1,\dots,d$; then the operator $\nabla^J$ is  $\nabla^J=(\partial_1^{J_1},\dots,\partial_d^{J_d})$. 
\item For $p\geq1$, $\|\cdot\|_p$ denotes the $p$-norm over $\R^d$, i.e., $\|x\|_p=\left(\sum_{i=1}^d|x_i|^p\right)^{\frac{1}{p}}$ for any $x\in\R^d$; $L^p_{loc}(\R^d)$ denotes the set of functions $f:\R^d\to\R$ such that $\int_\X |f(x)|^pdx<\infty$ for any compact subset $\X\subset\R^d$.
\item Let $J$ be a $d$-tuple multi-index of order $|J|$. For a function $f\in L^{1}_{loc}(\R^d)$, its $J^{th}$-weak derivative $D^Jf\in L^{1}_{loc}(\R^d)$ is a function such that for any smooth and compactly supported test function $g$, $\int_{\R^d}D^Jf(x)g(x)dx=(-1)^{|J|}\int_{\R^d}f(x)\nabla^Jg(x)dx$. The Sobolev space $W^{k,p}_{loc}(\R^d)$ is a set of functions $f$ on $\R^d$ such that for any $d$-tuple multi-index $J$ with $|J|\leq k$, $D^Jf\in L^p_{loc}(\R^d)$. 
\item Fix an arbitrary $\alpha\in\N^+$.  $G^\alpha(\R^d)$ denotes a subspace of $\CCal^\alpha(\R^d;\R)$ where for any $g\in G^\alpha(\R^d)$ and any multi-index $J$ with $|J|=\sum_{i=1}^dJ_i\leq \alpha$, there exist $k_1,k_2\in\N$ such that 
\[\nabla^Jg(x)\leq k_1\left(1+\|x\|_2^{2k_2}\right),\quad \forall x\in\R^d.\]
If $g$ is a parametrized functions $g_\beta$, then $g_\beta\in G^\alpha(\R^d)$ indicates that the choices of constants $k_1$ and $k_2$ are uniform over all possible $\beta$'s.
\item Fix an arbitrary $\alpha\in\N^+$. $G^\alpha_w(\R^d)$ denotes a subspace of $W^{\alpha,1}_{loc}(\R^d)$ where for any $g\in G^\alpha_w(\R^d)$ and any multi-index $J$ with $|J|=\sum_{i=1}^dJ_i\leq \alpha$, there exist $k_1,k_2\in\N$ such that 
\[D^Jg(x)\leq k_1\left(1+\|x\|_2^{2k_2}\right),\quad \text{for almost all } x\in\R^d.\]
If $g$ is a parametrized functions $g_\beta$, then $g_\beta\in G^\alpha_w(\R^d)$ indicates that the choices of constants $k_1$ and $k_2$ are uniform over all possible $\beta$'s.
\end{itemize}
\section{GANs training}\label{sec: train}
In this section, we will provide the mathematical setup for GANs training.
\subsection{GANs training: minimax versus maximin}
GANs fall into the category of generative models to approximate an unknown probability distribution $\pbm_r$. GANs are minimax games between two competing neural networks, the generator $G$ and the discriminator $D$. The neural network for the generator $G$ maps a latent random variable $Z$ with a known distribution $\pbm_z$ into the sample space to mimic the true distribution $\pbm_r$. Meanwhile, the other neural network for the discriminator $D$ will assign a score between $0$ to $1$ to an input sample, either a generated sample or a true one. A higher score from the discriminator $D$ indicates that the sample is more likely to be from the true distribution. 

Formally, let $(\Omega, \mathcal F, \{\mathcal{F}_t\}_{t\geq0}, \pbm)$ be a filtered probability space. Let a measurable space $\mathcal X\subset\R^{d_x}$ be the sample space. Let a $\X$-valued random variable $X$ denote the random sample, where $X:\Omega\to\X$ is a measurable function. The unknown probability distribution $\pbm_r=Law(X)$ such that $\pbm_r(X\in A)=\pbm(\{\omega\in\Omega:X(\omega)\in A\})$ for any measurable set $A\subset X$. Similarly, let a measurable space $\mathcal Z\subset \R^{d_z}$ be the latent space. Let a $\mathcal Z$-valued random variable $Z$ denote the latent variable where $Z:\Omega\to\mathcal Z$. The prior distribution $\pbm_z=Law(Z)$ such that $\pbm_z(Z\in B)=\pbm(\{\omega\in\Omega:Z(\omega)\in B\})$ for any measurable $B\subset \mathcal Z$. Moreover, $X$ and $Z$ are independent, i.e., \[\pbm\left(\{\omega:X(\omega)\in A,\,Z(\omega)\in B\}\right)=\pbm_r(X\in A)\pbm_z(Z\in B)\]
for any measurable sets $A\subset \mathcal X$ and $B\subset\mathcal Z$.

In the vanilla GANs framework by \cite{Goodfellow2014}, the loss function with respect to $G$ and $D$ is given by
\[L(G,D)=\E_{X\sim\pbm_r}\log D(X)+\E_{Y\sim G\#\pbm_z}[1-\log D(Y)],\]
and the objective is given by a {\em minimax} problem,
\[\min_G\max_D L(G,D).\]
Under a given $G$, the concavity of $L(G,D)$ with respect to $D$ follows from the concavity of functions $\log x$ and $\log(1-x)$; under a given $D$, the convexity of $L(G,D)$ with respect to $G$ follows from the linearity of pushforward measure $G\#\pbm_z$ and expectation. Therefore, the training loss in vanilla GANs is indeed convex in $G$ and concave in $D$. In the practical training stage, both $G$ and $D$ become parametrized neural networks $G_\theta$ and $D_\omega$ and therefore the working loss function is indeed with respect to the parameter $(\params)$,
\[\hat L(\params)=\E_{X\sim\pbm_r}\log D_\omega(X)+\E_{Z\sim \pbm_z}[1-\log D_\omega(G_\theta(Z))].\]
According to the training scheme proposed by \cite{Goodfellow2014}, in each iteration, $\omega$ is updated first followed by the update of $\theta$. It precisely corresponds to the {\em minimax} formulation of the objective,
\[\min_\theta\max_\omega \hat L(\params).\]
However in the practice training stage of GANs, there might be an interchange of training orders between the generator and the discriminator. One should be careful as the interchange implicitly modifies the objective into a {\em maximin} problem,
\[\max_\omega\min_\theta \hat L(\params),\]
and hence raises the question of whether these two objectives are equivalent. This question is closely related to the notion of Nash equilibrium in a two-player zero-sum game. According to the original GANs framework, the solution should provide an {\em upper value} to the corresponding two-player zero-sum game between the generator and the discriminator, i.e. an upper bound for the game value. As pointed out by Sion's theorem (see \cite{von1959theory} and \cite{sion1958general}), a sufficient condition to guarantee the equivalence between the two training orders is that the loss function $\hat L$ is concex in $\theta$ and concave in $\omega$. Though we have seen that the loss function $L$ with respect to $G$ and $D$ satisfies this condition, it is not necessarily true for $\hat L(\params)$. In fact, the work of \cite{zhu2020deconstructing} points out that these conditions are usually not satisfied with respect to generator and discriminator parameters in common GANs models and this lack of convexity and/or concavity does create challenges in the training of GANs. Such challenges motivate us to take a closer look at the evolution of parameters in the training of GANs using mathematical tools. In the following analysis, we will strictly follow the {\em minimax} formulation and its corresponding training order.

\subsection{SGA for GANs training}
Typically, GANs are trained through a stochastic gradient algorithm (SGA). An SGA is applied to a class of optimization problems whose loss function $\Phi(\gamma)$ with respect to model parameter vector $\gamma$ can be written as 
\[\Phi(\gamma)=\E_{\mathcal I}[\Phi_{\mathcal I}(\gamma)],\]
where a random variable $\ICal$ takes values in the index set $\II$ of the data points and for any $i\in\II$, $\Phi_i(\gamma)$ denotes the loss evaluated at the data point with index $i$. 

Suppose the objective is to minimize $\Phi(\gamma)$ over $\gamma$. Applying gradient descent with learning rate $\eta>0$, at an iteration $k$, $k=0,1,2,\dots$, the parameter vector is updated by
\[\gamma_{k+1}=\gamma_k-\eta\nabla\Phi(\gamma_k).\]
By linearity of differentiability and expectation, the above update can be written as
\[\gamma_{k+1}=\gamma_k-\eta\E_\ICal[\nabla\Phi_\ICal(\gamma_k)].\]
Under suitable conditions, $\E_\ICal[\nabla\Phi_\ICal(\gamma_k)]$ can be estimated by sample mean \[\hat\E_{\B}[\Phi_\ICal(\gamma)]=\frac{\sum_{k=1}^B\Phi_{I_k}(\gamma)}{B},\]
where $\B=\{I_1,\dots,I_B\}$ is a collection of indices with $I_k\overset{i.i.d.}{\sim}\ICal$, called a minibatch, and $B\ll|\II|$.

Under an SGA, the uncertainty in sampling $\B$ propagates through the training process, making it a stochastic process rather than a deterministic one. This stochasticity motivates us to study a continuous-time approximation for GANs training in the form of SDEs, as will be seen in \eqref{eq: dyn-approx} and \eqref{eq: gan-dyn}. (See also the connection between stochastic gradient descent and Markov chains in \citet{Dieuleveut2020}).

Consider the GANs training performed on a data set $\D=\{(z_i,x_j)\}_{1\leq i\leq N,\,1\leq j\leq M}$, where $\{z_i\}_{i=1}^N$ are sampled from $\pbm_z$ and $\{x_j\}_{j=1}^M$ are real image data following the unknown distribution $\pbm_r$. Let $\gene:\mathcal Z\to\X$ denote the generator parametrized by the neural network with the set of parameters $\theta\in\R^{d_\theta}$, and let $\disc:\X\to\R^+$ denote the discriminator parametrized by the other neural network with the set of parameters $\omega\in\R^{d_\omega}$, where $\R^+$ denotes the set of nonnegative real numbers. Then the objective of GANs is to solve the following minimax problem
\begin{equation}
    \label{eq: minimax-obj}
    \min_\theta\max_\omega\Phi(\theta,\omega),
\end{equation}
 for some cost function $\Phi$, with $\Phi$ of the form
 \begin{equation}
     \Phi(\params)=\frac{\sum_{i=1}^N\sum_{j=1}^MJ(\disc(x_j),\disc(\gene(z_i)))}{N\cdot M}.
 \end{equation}
For instance, $\Phi$ in the vanilla GANs model \citet{Goodfellow2014} is given by
\begin{equation*}
    \begin{aligned}
        \Phi(\theta, \omega)
        &=\frac{\sum_{i=1}^N\sum_{j=1}^M\log \disc(x_j)+\log(1-\disc(\gene(z_i)))}{N\cdot M},
    \end{aligned}
\end{equation*}
while $\Phi$ in Wasserstein GANs \citet{Arjovsky2017} takes the form
\begin{equation*}
    \begin{aligned}
        \Phi(\theta, \omega)
        &=\frac{\sum_{i=1}^N\sum_{j=1}^M \disc(x_j)-\disc(\gene(z_i))}{N\cdot M}.
    \end{aligned}
\end{equation*}
 Here the full gradients of $\Phi$ with respect to $\theta$ and $\omega$ are estimated over a mini-batch $\B$ of batch size $B$. One way of sampling $\B$ is to choose $B$ samples out of a total of $N\cdot M$ samples without putting back, another is to take $B$ i.i.d. samples. The analyses for both cases are similar, here we adopt the second sampling scheme. 
 
 More precisely, let  $\B=\{(z_{I_k},x_{J_k})\}_{k=1}^B$ be  i.i.d. samples from $\D$. Let $\gt$ and $\go$ be the full gradients of $\Phi$ with respect to $\theta$ and $\omega$ such that
\begin{equation}\label{eq: def-full-grads}
\begin{aligned}
  &\gt(\params)=\nabla_\theta\Phi(\theta, \omega)=\frac{\sum_{i=1}^N\sum_{j=1}^M\gtij(\params)}{N\cdot M},\\
  &\go(\theta,\omega)=\nabla_\omega\Phi(\theta,\omega)=\frac{\sum_{i=1}^N\sum_{j=1}^M\goij(\params)}{N\cdot M}.
\end{aligned}
\end{equation}
Here $\gtij$ and $\goij$ denote $\nabla_\theta J(D_\omega(x_j),D_\omega(G_\theta(z_i)))$ and $\nabla_\omega J(D_\omega(x_j),D_\omega(G_\theta(z_i)))$, respectively, with differential operators defined as
\[\nabla_\theta:=\begin{pmatrix}\partial_{\theta_1}&\cdots& \partial_{\theta_{d_\theta}}\end{pmatrix}^T,\quad
\nabla_\omega:=\begin{pmatrix}\partial_{\omega_1}& \cdots&\partial_{\omega_{d_\omega}}\end{pmatrix}^T.\] 
Then, the estimated gradients for $\gt$ and $\go$ corresponding to the mini-batch $\mathcal B$ are
\begin{equation}
\begin{aligned}
  &\gtb(\params)=\frac{\sum_{k=1}^B\gtbk(\params)}{B},\quad \gob(\params)=\frac{\sum_{k=1}^B\gobk(\params)}{B}.
\end{aligned}
\end{equation}

Moreover, let $\eta^\theta_t>0$ and $\eta^\omega_t>0$ be the learning rates at iteration $t=0,1,2,\dots$, for $\theta$ and $\omega$ respectively, then solving the minimax problem \eqref{eq: minimax-obj} with SGA and {\it alternating parameter update} implies descent of $\theta$ along $\gt$ and ascent of $\omega$ along $\go$ at each iteration, i.e., 
\begin{equation}
    \label{eq: alt-update}
\left\{\begin{aligned}
      \omega_{t+1}&=\omega_t+\lro_t\gob(\theta_t,\omega_t),\\
      \theta_{t+1}&=\theta_t-\lrt_t\gtb(\theta_t,\omega_{t+1}).
    \end{aligned}\right.
\end{equation}
Furthermore, within each iteration, the minibatch gradient for $\theta$ and $\omega$ are calculated on different batches. In order to emphasize this difference, we use $\bar\B$ to represent the minibatch for $\theta$ and $\B$ for that of $\omega$, with $\bar\B\overset{i.i.d.}{\sim}\B$. That is,
\begin{equation}
    \label{eq: dt-gan-update}\tag{ALT}
    \left\{\begin{aligned}
      &\omega_{t+1}=\omega_t+\lro_t\gob(\theta_t,\omega_t),\\
      &\theta_{t+1}=\theta_t-\lrt_t\gt^{\bar\B}(\theta_t,\omega_{t+1}).
    \end{aligned}\right.
\end{equation}
Some practical training of GANs uses {\it simultaneous parameter update} between the discriminator and the generator, corresponding to a similar yet subtly different form
\begin{equation}
    \label{eq: sml-update}\tag{SML}
    \left\{\begin{aligned}
      &\omega_{t+1}=\omega_t+\lro_t\gob(\theta_t,\omega_t),\\
      &\theta_{t+1}=\theta_t-\lrt_t\gtb(\theta_t,\omega_{t}).
    \end{aligned}\right.
\end{equation}

For the ease of exposition, we will assume throughout the paper, an constant learning rates $\lrt_t=\lro_t=\eta$, with $\eta$ viewed as the time interval between two consecutive parameter updates. 
 
\section{Approximation and error bound analysis of GANs training}\label{sec: setup}
The randomness in sampling $\B$ (and $\bar\B$) makes the GANs training process prescribed by \eqref{eq: dt-gan-update} and \eqref{eq: sml-update} stochastic processes. In this section, we will establish their continuous-time approximations and error bounds, where the approximations are in the form of coupled SDEs.

\subsection{Approximation}
To get an intuition of how the exact expression of SDEs emerges, let us start by some basic properties embedded in the training process. First, let $I:\Omega\to\{1,\dots,N\}$ and $J:\Omega\to\{1,\dots,M\}$ denote random indices independently and uniformly distributed respectively, then according to the definitions of $\gt$ and $\go$ in \eqref{eq: def-full-grads}, we have 
$\E[\gtr(\params)]=\gt(\params)$ and $\E[\gor(\params)]=\go(\params)$.
Denote the correspondence covariance matrices as
\[\begin{aligned}
  &\sigt(\params)=\frac{\sum_i\sum_j[\gtij(\params)-\gt(\params)][\gtij(\params)-\gt(\params)]^T}{N\cdot M},\\
  &\sigo(\params)=\frac{\sum_i\sum_j[\goij(\params)-\go(\params)][\goij(\params)-\go(\params)]^T}{N\cdot M}, 
\end{aligned}\]
since $(I_k,J_k)$ in $\B$ are i.i.d. copies of $(I,J)$, then
\[\begin{aligned}
  &\E_\B[\gtb(\params)]=\E\left[\frac{\sum_{k=1}^B\gtbk(\params)}{B}\right]= \gt(\theta,\omega),\\
  &\E_\B[\gob(\params)]=\E\left[\frac{\sum_{k=1}^B\gobk(\params)}{B}\right]= \go(\theta,\omega),\\
  &Var_\B(\gtb(\params))=Var_\B\left(\frac{\sum_{k=1}^B\gtbk(\params)}{B}\right)=\frac{1}{B}\sigt(\params),\\
  &Var_\B(\gob(\params))=Var_\B\left(\frac{\sum_{k=1}^B\gobk(\params)}{B}\right)=\frac{1}{B}\sigo(\params),
\end{aligned}\]
As the batch size $B$ gets sufficiently large, the classical central limit theorem leads to the following approximation of \eqref{eq: dt-gan-update},
\begin{equation}\label{eq: gan-evol}
\left\{\begin{aligned}
  \omega_{t+1}&=\omega_t+\eta\gob(\theta_t,\omega_t)\approx \omega_t+\eta\go(\theta_t,\omega_t)+\frac{\eta}{\sqrt{B}}\sigo^{\frac{1}{2}}(\theta_t,\omega_t)Z^1_t,\\
  \theta_{t+1}&=\theta_t-\eta\gtb(\theta_{t},\omega_{t+1})\approx \theta_t-\eta\gt(\theta_t,\omega_{t+1})+\frac{\eta}{\sqrt{B}}\sigt^{\frac{1}{2}}(\theta_t,\omega_{t+1})Z^2_t,
\end{aligned}\right.
\end{equation}
with independent Gaussian random variables $Z^1_t{\sim} N(0,1\cdot I_{d_\omega})$ and $Z^2_t\sim N(0,1\cdot I_{d_\theta})$, $t=0,1,2,\dots$.
Here, the scalar 1 specifies the time increment $1=\Delta t=(t+1)-t$.

Write $t+1=t+\Delta t$. On one hand, assuming the continuity of the process $\{\omega_t\}_t$ with respect to time $t$ and sending $\Delta t$ to 0, one intuitive approximation in the following form can be easily derived,
 \begin{equation}
    \label{eq: pre-dyn-approx}
    \begin{aligned}
    d\begin{pmatrix}\Theta_t\\\mathcal{W}_t\end{pmatrix}=\begin{pmatrix}-\gt(\Theta_t,\W_t)\\\go(\Theta_t,\W_t)\end{pmatrix}dt
    +\sqrt{2\beta^{-1}}\begin{pmatrix}\sigt(\Theta_t,\W_t)^{\frac{1}{2}}&0\\
    0&\sigo(\Theta_t,\W_t)^{\frac{1}{2}}\end{pmatrix}dW_t,
    \end{aligned}
\end{equation}
with $\beta=\frac{2B}{\eta}$ and $\{W_t\}_{t\geq0}$ be standard $(d_\theta+d_\omega)$-dimensional Brownian motion supported by the filtered probability space $(\Omega, \mathcal F, \{\mathcal F_t\}_{t\geq0},\pbm)$. Let $\{\mathcal F^W_t\}_{t\geq0}$ denote the natural filtration generated by $\{W_t\}_{t\geq0}$. As an continuous-time approximation for GANs training, SDEs in this rather intuitive form are adopted without justification in some earlier works such as \citet{Domingo-Enrich2020} and \citet{Conforti2020}. Later we will show that \eqref{eq: pre-dyn-approx} is in fact an approximation for GANs training of \eqref{eq: sml-update}. 

On the other hand, the game nature in GANs is demonstrated through the interactions between the generator and the discriminator during the training process, especially the appearance of $\omega_{t+1}$ at the update of $\theta$ as in \eqref{eq: dt-gan-update}. However, the adapted coupled processes given by \eqref{eq: pre-dyn-approx} do not capture such interactions. One possible approximation for the GANs training process of \eqref{eq: dt-gan-update} would be 
  \begin{equation}
    \label{eq: pre-gan-dyn}
    \begin{aligned}
    d\begin{pmatrix}\Theta_t\\\mathcal{W}_t\end{pmatrix}&=\biggl[\begin{pmatrix}-\gt(\Theta_t,\mathcal{W}_t)\\\go(\Theta_t,\W_t)\end{pmatrix}+\frac{\eta}{2}\begin{pmatrix}\nabla_\theta\gt(\Theta_t,\mathcal{W}_t)&-\nabla_\omega\gt(\Theta_t,\mathcal{W}_t)\\-\nabla_\theta\go(\Theta_t,\mathcal{W}_t)&-\nabla_\omega\go(\Theta_t,\mathcal{W}_t)\end{pmatrix}\begin{pmatrix}-\gt(\Theta_t,\mathcal{W}_t)\\\go(\Theta_t,\mathcal{W}_t)\end{pmatrix}\biggl]dt\\
    &\hspace{30pt}+\sqrt{2\beta^{-1}}\begin{pmatrix}\sigt(\Theta_t,\W_t)^{\frac{1}{2}}&0\\
    0&\sigo(\Theta_t,\W_t)^{\frac{1}{2}}\end{pmatrix}dW_t.
    \end{aligned}
\end{equation}
Equations \eqref{eq: pre-dyn-approx} and \eqref{eq: pre-gan-dyn} can be written in more compact forms
\begin{align}d\begin{pmatrix}\Theta_t\\\mathcal{W}_t\end{pmatrix}&=b_0(\Theta_t,\W_t)dt+\sigma(\Theta_t,\W_t)dW_t,\label{eq: dyn-approx}\tag{SML-SDE}\\ d\begin{pmatrix}\Theta_t\\\mathcal{W}_t\end{pmatrix}&=b(\Theta_t,\W_t)dt+\sigma(\Theta_t,\W_t)dW_t.\label{eq: gan-dyn}\tag{ALT-SDE}\end{align}
where the drift $b(\params)=b_0(\params)+\eta b_1(\params)$, with
\begin{align}
    b_0(\params)&=\begin{pmatrix}-\gt(\params)\\\go(\params)\end{pmatrix},\\
    b_1(\params)&=\frac{1}{2}\begin{pmatrix}\nabla_\theta\gt(\params)&-\nabla_\omega\gt(\params)\\-\nabla_\theta\go(\params)&-\nabla_\omega\go(\params)\end{pmatrix}\begin{pmatrix}-\gt(\params)\\\go(\params)\end{pmatrix}\nonumber\\
    &=-\frac{1}{2}\nabla b_0(\params)b_0(\params)-\begin{pmatrix}\nabla_\omega\gt(\params)\go(\params)\\0\end{pmatrix},\label{eq: additional-term}\\
\text{and volatility}\hspace{5pt}
    \sigma(\params)&=\sqrt{2\beta^{-1}}\begin{pmatrix}\sigt(\Theta_t,\W_t)^{\frac{1}{2}}&0\\
    0&\sigo(\Theta_t,\W_t)^{\frac{1}{2}}\end{pmatrix}.\label{eq: volatility}
\end{align}
The drift terms in the SDEs, i.e., $b_0$ in \eqref{eq: dyn-approx} and $b$ in \eqref{eq: gan-dyn}, show the direction of the parameters evolution; the diffusion terms $\sigma$ represent the fluctuations of the learning curves for these parameters. Moreover, the form of SDEs prescribes $\beta$ the ratio between the batch size and the learning rate in order to modulate the fluctuations of SGAs in GANs training.
Even though both \eqref{eq: dyn-approx} and \eqref{eq: gan-dyn} are adapted to $\{\mathcal F^W_t\}_{t\geq0}$, the term 
\[-\frac{\eta}{2}\begin{pmatrix}\nabla_\omega\gt(\params)\go(\params)\\0\end{pmatrix}\] 
in \eqref{eq: gan-dyn} highlights the interaction between the generator and the discriminator in GANs training process; see Remark \ref{rmk: game}.

\subsection{Error bound for the SDE approximation}
We will show that these coupled SDEs are indeed the continuous-time approximations of GANs training processes, with precise error bound analysis. 
Here the approximations are under the notion of weak approximation as in \citet{Li2019}. 
More precisely, the following Theorems \ref{thm: gan-approx} and \ref{thm: sml-sde} provide conditions under which the evolution of parameters in GANs are within a reasonable distance from its SDE approximation.

\begin{theorem}\label{thm: gan-approx}
Fix an arbitrary time horizon $\mathcal T>0$ and take the learning rate $\eta\in(0,1\wedge \mathcal T)$ and the number of iterations $N=\left\lfloor\frac{\mathcal T}{\eta}\right\rfloor$. Suppose that
\begin{enumerate}[label=\ref{thm: gan-approx}.\alph*]
    \item $\goij$ is twice continuously differentiable, and $\gtij$ and $\goij$ are Lipschitz, for any $i=1,\dots,N$ and $j=1,\dots,M$;
    \item $\Phi$ is of $\mathcal{C}^3(\R^{d_\theta+d_\omega})$ and $\Phi\in G^{4}_{w}(\R^{d_\theta+d_\omega})$;
    \item $(\nabla_\theta\gt)\gt$, $(\nabla_\omega\gt)\go$, $(\nabla_\theta\go)\gt$ and $(\nabla_\omega\go)\go$ are all Lipschitz.
\end{enumerate}
Then, $(\Theta_{t\eta},\mathcal W_{t\eta})$ as in \eqref{eq: gan-dyn} is a weak approximation of $(\theta_t,\omega_t)$ as in \eqref{eq: dt-gan-update} of order 2, i.e., given any initialization  $\theta_0=\theta$ and $\omega_0=\omega$, for any test function $f\in G^3(\R^{d_\theta+d_\omega})$,
we have the following estimate
\begin{equation}
\label{eq: alt-error}
\max_{t=1,\dots,N}\left|\E f(\theta_t,\omega_t)-\E f(\Theta_{t\eta},\mathcal W_{t\eta})\right|\leq C\eta^2
\end{equation}
for constant $C\geq0$.
\end{theorem}

\begin{theorem}
\label{thm: sml-sde}
Fix an arbitrary time horizon $\mathcal T>0$, take the learning rate $\eta\in(0,1\wedge \mathcal T)$ and the number of iterations $N=\left\lfloor\frac{\mathcal T}{\eta}\right\rfloor$. Suppose
\begin{enumerate}[label=\ref{thm: sml-sde}.\alph*]
    \item $\Phi(\theta,\omega)$ is continuously differentiable and $\Phi\in G^{3,1}_{W}(\R^{d_\theta+d_\omega})$; 
    \item $\gtij$ and $\goij$ are Lipschitz for any $i=1,\dots, N$ and $j=1,\dots,M$.
\end{enumerate}
Then, $(\Theta_{t\eta},\mathcal W_{t\eta})$ as in \eqref{eq: dyn-approx} is a weak approximation of $(\theta_t,\omega_t)$ as in \eqref{eq: sml-update} of order 1, i.e., given any initialization $\theta_0=\theta$ and $\omega_0=\omega$, for any test function $f\in G^2(\R^{d_\theta+d_\omega})$,
we have the following estimate
\begin{equation}\label{eq: sml-error}
\max_{t=1,\dots,N}\left|\E f(\theta_t,\omega_t)-\E f(\Theta_{t\eta},\mathcal W_{t\eta})\right|\leq C\eta
\end{equation} for constant $C\geq0$.
\end{theorem}
These approximations in Theorems \ref{thm: gan-approx} and \ref{thm: sml-sde} will enable us to analyze the long-run behavior of GANs training later in Section \ref{sec: conv}, via studying the invariant measures of SDEs.

\begin{remark}\label{rmk: game}
Modifying the intuitive SDE approximation \eqref{eq: dyn-approx} into
\begin{equation}\label{eq: tighter-sml}
d\begin{pmatrix}\Theta_t\\\W_t\end{pmatrix}=\biggl[b_0(\Theta_t,\W_t)-\frac{\eta}{2}\nabla b_0(\Theta_t,\W_t)b_0(\Theta_t,\W_t)\biggl]dt+\sigma(\Theta_t,\W_t)dW_t
\end{equation}
and applying similar techniques in proving Theorem \ref{thm: gan-approx}, one can get an $O(\eta^2)$ approximation for \eqref{eq: sml-update}. However, comparing \eqref{eq: tighter-sml} and \eqref{eq: gan-dyn}, the term
\[-\frac{\eta}{2}\begin{pmatrix}\nabla_\omega\gt(\params)\go(\params)\\0\end{pmatrix}\]
still stands out, which is due to the interactions between the generator and discriminator during training. It implies that the ``game effect'' between the generator and the discriminator has an impact to the evolution trajectories of the model parameters.
\end{remark}

\subsection{Proofs of Theorem \ref{thm: gan-approx}}\label{subsec: thms3.1-3.2}
In this section we will provide a detailed proof of Theorem \ref{thm: gan-approx}; proof of Theorem \ref{thm: sml-sde} is a simple analogy and thus omitted. We will adapt the approach from \citet{Li2019} to our analysis of GANs training.

\subsubsection{Preliminary analysis}
\paragraph{One-step difference.}
Recall that under the alternating update scheme and constant learning rate $\eta$, the GANs training is as follows,
\begin{equation}
    \label{eq: dt-gan-update}\tag{ALT}
    \left\{\begin{aligned}
      &\omega_{t+1}=\omega_t+\eta\gob(\theta_t,\omega_t),\\
      &\theta_{t+1}=\theta_t-\eta\gt^{\bar\B}(\theta_t,\omega_{t+1}),
    \end{aligned}\right.
\end{equation}
where $\B$ and $\bar\B$ are i.i.d., emphasizing the fact that the evaluations of gradients are performed on different mini-batches when updating $\theta$ and $\omega$ alternatively. 

Let $(\theta,\omega)$ denote the initial value for $(\theta_0,\omega_0)$ and
\begin{equation}\label{eq: alt-diff}
\Delta=\Delta(\theta,\omega)=\begin{pmatrix}\theta_1-\theta\\\omega_1-\omega\end{pmatrix}
\end{equation}
be the one-step difference. Let $\Delta^{i,j}$ denote the tuple consisting of the $i$-th and $j$-th component of one-step difference of $\theta$ and $\omega$, respectively, with $i=1,\dots,d_{\theta}$ and $j=1,\dots,d_{\omega}$. 

\begin{lemma}\label{lem: alt-diff}
Assume that $\gtij$ is twice continuously differentiable for any $i=1,\dots, N$ and $j=1,\dots, M$. 
\begin{enumerate}
    \item The first moment is given by
    \[\E[\Delta^{i,j}]=\eta\begin{pmatrix}-\gt(\theta,\omega)_{i}\\\go(\theta,\omega)_j\end{pmatrix}+\eta^2\begin{pmatrix}\left\{-\nabla_\omega[\gt(\theta,\omega)_i]\right\}^T\go(\theta,\omega)\\0\end{pmatrix}+O(\eta^3).\]
    \item The second moment is given by
    \[\begin{aligned}
        \E[\Delta^{i,j}(\Delta^{k,l})^T]=&\eta^2\left[\frac{1}{B}\begin{pmatrix}\sigt(\theta,\omega)_{i,k}&0\\
        0&\sigo(\theta,\omega)_{j,l}\end{pmatrix}+\begin{pmatrix}-\gt(\theta,\omega)_{i}\\\go(\theta,\omega)_j\end{pmatrix}\begin{pmatrix}-\gt(\theta,\omega)_{k}\\\go(\theta,\omega)_l\end{pmatrix}^T\right]\\
        &+O(\eta^3),
    \end{aligned}\]
    where $\sigt(\theta,\omega)_{i,k}$ and $\sigo(\theta,\omega)_{j,l}$ denote the element at position $(i,k)$ and $(j,l)$ of matrices $\sigt(\params)$ and $\sigo(\params)$, respectively.
    \item The third moments are all of order $O(\eta^3)$.
\end{enumerate}
\end{lemma}
\begin{proof} By a second-order Taylor expansion, we have
\begin{equation}
    \label{eq: est-delta}
    \Delta(\theta,\omega)=\eta\begin{pmatrix}-\gt^{\bar\B}(\theta,\omega)\\\gob(\theta,\omega)\end{pmatrix}+\eta^2\begin{pmatrix}-\nabla_\omega\gt^{\bar\B}(\theta,\omega))\gob(\theta,\omega)\\0\end{pmatrix}+O(\eta^3).
\end{equation}
Then,
\begin{align}
  &\Delta^{i,j}(\params)=\eta\begin{pmatrix}-\gt^{\bar\B}(\theta,\omega)_{i}\\\gob(\theta,\omega)_j\end{pmatrix}+\eta^2\begin{pmatrix}\left\{-\nabla_\omega[\gt^{\bar\B}(\theta,\omega)_i]\right\}^T\gob(\theta,\omega)\\0\end{pmatrix}+O(\eta^3),\\
  &\Delta^{i,j}(\params)[\Delta^{k,l}(\params)]^T=\eta^2\begin{pmatrix}\gt^{\bar\B}(\params)_i\gt^{\bar\B}(\params)_k & -\gt^{\bar\B}(\params)_i\gob(\params)_l\\
  -\gt^{\bar\B}(\params)_k\gob(\params)_j & \gob(\params)_j\gob(\params)_l\end{pmatrix}+O(\eta^3),
\end{align}
and higher order polynomials are of order $O(\eta^3)$. Notice that $\bar\B\perp\B$ and recall the definition of $\sigt$ and $\sigo$. The conclusion follows.\hfill$\square$
\end{proof}

Now consider the following SDE,
\begin{equation}
d\begin{pmatrix}\Theta_t\\\mathcal{W}_t\end{pmatrix}=b(\Theta_t,\W_t)dt+\sigma(\Theta_t,\W_t)dW_t,\label{eq: gan-dyn}\tag{ALT-SDE}
\end{equation}
where $b(\params)=b_0(\params)+\eta b_1(\params)$, with
\begin{align}
    b_0(\params)&=\begin{pmatrix}-\gt(\params)\\\go(\params)\end{pmatrix},\label{eq: b0}\\
    b_1(\params)&=\frac{1}{2}\begin{pmatrix}\nabla_\theta\gt(\params)&-\nabla_\omega\gt(\params)\\-\nabla_\theta\go(\params)&-\nabla_\omega\go(\params)\end{pmatrix}\begin{pmatrix}-\gt(\params)\\\go(\params)\end{pmatrix}\nonumber\\
    &=-\frac{1}{2}\nabla b_0(\params)b_0(\params)-\begin{pmatrix}\nabla_\omega\gt(\params)\go(\params)\\0\end{pmatrix},\label{eq: additional-term}\\
\text{and}\hspace{5pt}
    \sigma(\params)&=\sqrt{2\beta^{-1}}\begin{pmatrix}\sigt(\Theta_t,\W_t)^{\frac{1}{2}}&0\\
    0&\sigo(\Theta_t,\W_t)^{\frac{1}{2}}\end{pmatrix}.\label{eq: sig0}
\end{align}
With the same initialization like \eqref{eq: alt-diff}, define the corresponding one-step difference for \eqref{eq: gan-dyn},
\begin{equation}\label{eq: alt-sde-diff}
\tilde\Delta=\tilde\Delta(\params)=\begin{pmatrix}\Theta_{1\times\eta}-\theta\\\W_{1\times\eta}-\omega\end{pmatrix}.
\end{equation}
Let $\tilde\Delta_k$ be the $k$-th component of $\tilde\Delta$, $k=1,\dots, d_\theta+d_\omega$ and $\tilde\Delta^{i,j}$ be the tuple consisting of the $i$-th and $j$-th component of one-step difference of $\Theta$ and $\W$, respectively, with $i=1,\dots,d_{\theta}$ and $j=1,\dots,d_{\omega}$.
\begin{lemma}\label{lem: alt-sde-diff}
Suppose $b_0$, $b_1$ and $\sigma$, given by \eqref{eq: b0},\eqref{eq: additional-term} and \eqref{eq: sig0}, are from $\mathcal{C}^3(\R^{d_\theta+d_\omega})$ such that for any multi-index $J$ of order $|J|\leq 3$, there exist $k_1, k_2\in\mathbb N$ satisfying
\[\max\{|\nabla^Jb_0(\params)|,|\nabla^Jb_1(\params)|,|\nabla^J\sigma(\params)|\}\leq k_1\biggl(1+\biggl\|\begin{pmatrix}\theta\\\omega\end{pmatrix}\biggl\|_2^{2k_2}\biggl)\]
and they are all Lipschitz. Then
\begin{enumerate}
    \item The first moment is given by
    \[\E[\tilde\Delta^{i,j}]=\eta\begin{pmatrix}-\gt(\theta,\omega)_{i}\\\go(\theta,\omega)_j\end{pmatrix}+\eta^2\begin{pmatrix}\left\{-\nabla_\omega[\gt(\theta,\omega)_i]\right\}^T\go(\theta,\omega)\\0\end{pmatrix}+O(\eta^3).\]
    \item The second moment is given by
    \[\begin{aligned}
        \E[\tilde\Delta^{i,j}(\tilde\Delta^{k,l})^T]=&\eta^2\left[\frac{1}{B}\begin{pmatrix}\sigt(\theta,\omega)_{i,k}&0\\
        0&\sigo(\theta,\omega)_{j,l}\end{pmatrix}+\begin{pmatrix}-\gt(\theta,\omega)_{i}\\\go(\theta,\omega)_j\end{pmatrix}\begin{pmatrix}-\gt(\theta,\omega)_{k}\\\go(\theta,\omega)_l\end{pmatrix}^T\right]\\
        &+O(\eta^3).
    \end{aligned}\]
    \item The third moments are all of order $O(\eta^3)$.
\end{enumerate}
\end{lemma}
\begin{proof}Let $\psi:\R^{d_\theta+d_\omega}\to\R$ be any smooth test function. Under the dynamic \eqref{eq: gan-dyn}, define the following operators
\begin{align*}
&\LL_1\psi(\params)=b_0(\params)^T\nabla\psi(\params),\\
&\LL_2\psi(\params)=b_1(\params)^T\nabla\psi(\params),\\
&\LL_3\psi(\params)=\frac{1}{2}Tr\biggl(\sigma(\params)\sigma(\params)^T\nabla^2\psi(\params)\biggl).
\end{align*}
Apply It\^o's formula to $\psi(\Theta_t,\W_t)$, $\LL_i\psi(\Theta_t,\W_t)$ for $i=1,2,3$, and $\LL_1^2\psi(\Theta_t,\W_t)$, we have 
\begin{align}
&\psi(\Theta_\eta,\W_\eta)=\psi(\params)+\int_0^\eta(\LL_1+\eta\LL_2+\LL_3)\psi(\Theta_t,\W_t)dt+\int_0^\eta[\nabla\psi(\Theta_t,\W_t)]^T\sigma(\Theta_t,\W_t)dW_t\nonumber\\
&\hspace{5pt}=\psi(\params)+\eta\biggl(\LL_1+\LL_3\biggl)\psi(\params) + \eta^2\biggl(\frac{1}{2}\LL_1^2+\LL_2\biggl)\psi(\params)\label{eq: mean}\\
&\hspace{5pt}\left.\begin{aligned}&+\int_0^\eta\int_0^t\int_0^s\LL_1^3\psi(\Theta_u,\W_u)dudsdt+\int_0^\eta\int_0^t\biggl(\LL_3\LL_1+\LL_1\LL_3+\LL_3^2\biggl)\psi(\Theta_s,\W_s)dsdt\\
&+\eta\int_0^\eta\int_0^t\biggl(\LL_2\LL_1+\LL_1\LL_2+\LL_3\LL_2+\LL_2\LL_3\biggl)\psi(\Theta_s,\W_s)dsdt\\
&+\eta^2\int_0^\eta\int_0^t\LL_2^2\psi(\Theta_s,\W_s)dsdt \end{aligned}\right\}\label{eq: eta-cube}\\
&\hspace{5pt}+ M_\eta,
\end{align}
where $M_\eta$ denotes the remaining martingale term with mean zero. Given the regularity conditions of $b_0$, $b_1$ and $\sigma$, \cite[Theorem 9 in Section 2.5]{Krylov2008} implies that \eqref{eq: eta-cube} is of order $O(\eta^3)$. Therefore, 
\[\E\biggl[\psi(\Theta_\eta,\W_\eta)\biggl|\Theta_0=\theta,\W_0=\omega\biggl]=\psi(\params)+\eta\biggl(\LL_1+\LL_3\biggl)\psi(\params) + \eta^2\biggl(\frac{1}{2}\LL_1^2+\LL_2\biggl)\psi(\params).\]
Take $\psi(\Theta_\eta,\W_\eta)$ as $\tilde\Delta_i$, $\tilde\Delta_i\tilde\Delta_j$ and $\tilde\Delta_i\tilde\Delta_j\tilde\Delta_k$ for arbitrary indices $i,j,k=1,\dots,d_\theta+d_\omega$, then the conclusion follows.\hfill$\square$
\end{proof}

\paragraph{Estimate of moments.}
Next, we will bound the moments of GANs parameters under \eqref{eq: dt-gan-update}.
\begin{lemma}
\label{lem: moments}
Fix an arbitrary time horizon $\mathcal T>0$ and take the learning rate $\eta\in(0,1\wedge \mathcal T)$ and the number of iterations $N=\left\lfloor\frac{\mathcal T}{\eta}\right\rfloor$. Suppose that $\gtij$ and $\goij$ are all lipschitz, i.e. there exists $L>0$ such that 
\[\max_{i,j}\{|\gtij(\params)|,|\goij(\params)|\}\leq L\biggl(1+\biggl\|\begin{pmatrix}\theta\\\omega\end{pmatrix}\biggl\|_2\biggl).\]
Then for any $m\in\mathbb N$, $\max_{t=1,\dots,N}\E\biggl[\biggl\|\begin{pmatrix}\theta_t\\\omega_t\end{pmatrix}\biggl\|_2^m\biggl]$ is uniformly bounded, independent from $\eta$.
\end{lemma}
\begin{proof}
Throughout the proof, positive constants $C$ and $C'$ may vary from line to line. The Lipschitz assumption suggests that
\[\max\{|\gtb(\params)|, |\gob(\params)|\}\leq L\biggl(1+\biggl\|\begin{pmatrix}\theta\\\omega\end{pmatrix}\biggl\|_2\biggl).\]
For any $k=1,\dots,m$
\[\max\{|\gtb(\params)|^k, |\gob(\params)|^k\}\leq L\cdot k\binom{k}{\lfloor\frac{k}{2}\rfloor}\cdot\biggl(1+\biggl\|\begin{pmatrix}\theta\\\omega\end{pmatrix}\biggl\|_2^k\biggl),\]
and 
\[\biggl\|\begin{pmatrix}\theta\\\omega\end{pmatrix}\biggl\|_2^k+\biggl\|\begin{pmatrix}\theta\\\omega\end{pmatrix}\biggl\|_2^m\leq 2\biggl(1+\biggl\|\begin{pmatrix}\theta\\\omega\end{pmatrix}\biggl\|_2^m\biggl).\]
For any $t=0,\dots, N-1$,
\begin{align*}
\biggl\|\begin{pmatrix}\theta_{t+1}\\\omega_{t+1}\end{pmatrix}\biggl\|_2^m&\leq \biggl\|\begin{pmatrix}\theta_t\\\omega_t\end{pmatrix}\biggl\|_2^m+\sum_{k=1}^m\binom{m}{k}\biggl\|\begin{pmatrix}\theta_t\\\omega_t\end{pmatrix}\biggl\|_2^{m-k}\eta^k\biggl\|\begin{pmatrix}-\gtb(\theta_t,\omega_t)\\\gob(\theta_t,\omega_t)\end{pmatrix}\biggl\|_2^k\\
&\leq \biggl\|\begin{pmatrix}\theta_t\\\omega_t\end{pmatrix}\biggl\|_2^m+C\eta\sum_{k=1}^m\binom{m}{k}\biggl\|\begin{pmatrix}\theta_t\\\omega_t\end{pmatrix}\biggl\|_2^{m-k}\biggl(1+\biggl\|\begin{pmatrix}\theta_t\\\omega_t\end{pmatrix}\biggl\|_2^m\biggl)\\
&\leq \biggl(1+C\eta\biggl)\biggl\|\begin{pmatrix}\theta_t\\\omega_t\end{pmatrix}\biggl\|_2^m+C'\eta.
\end{align*}
Denote $a^m_t=\biggl\|\begin{pmatrix}\theta_t\\\omega_t\end{pmatrix}\biggl\|_2^m$. Then, $a^m_{t+1}\leq (1+C\eta)a^m_t+C'\eta$ that leads to
\begin{align*}
a^m_t&\leq (1+C\eta)^t\biggl(a^m_0+\frac{C'}{C}\biggl)-\frac{C'}{C}\\
&\leq (1+C\eta)^{\frac{\mathcal T}{\eta}}\biggl(a^m_0+\frac{C'}{C}\biggl)-\frac{C'}{C}\\
&\leq e^{C\mathcal T}\biggl(a^m_0+\frac{C'}{C}\biggl)-\frac{C'}{C}.
\end{align*}
The conclusion follows.
\hfill$\square$
\end{proof}

\paragraph{Mollification.}
Notice that in Theorem \ref{thm: gan-approx} (and Theorem \ref{thm: sml-sde}), the condition about the differentiability of loss function $\Phi$ is in the weak sense. For the ease of analysis, we will adopt the following mollification, given in \citet{Evans1998}.
\begin{definition}[Mollifier]
Define the following function $\nu:\R^{d_\theta+d_\omega}\to\R$,
\begin{equation*}
  \nu(u)=\begin{cases}
    C\exp\biggl\{-\frac{1}{\|u\|_2^2-1}\biggl\},&\|u\|_2<1;\\
    0,&\|u\|_2\geq1,
  \end{cases}
\end{equation*}
such that $\int_{\R^{d_\theta+d_\omega}}\nu(u)du=1$. For any $\epsilon>0$, define $\nu^\epsilon(u)=\frac{1}{\epsilon^{d_\theta+d_\omega}}\nu\left(\frac{u}{\epsilon}\right)$.
\end{definition}
Note that the mollifier $\nu\in\mathcal C^{\infty}(\R^{d_\theta+d_\omega})$ and for any $\epsilon>0$, $supp(\nu^{\epsilon})=B_\epsilon(0)$ where $B_\epsilon(0)$ denotes the $\epsilon$ ball around the origin in the Euclidean space $\R^{d_\theta+d_\omega}$.
\begin{definition}[Mollification]
Let $f\in\mathcal L_{loc}^1(\R^{d_\theta+d_\omega})$ be any locally integrable function. For any $\epsilon>0$, define $f^\epsilon=\nu^\epsilon*f$ such that
\[f^\epsilon(u)=\int_{\R^{d_\theta+d_\omega}}\nu^\epsilon(u-v)f(v)dv=\int_{\R^{d_\theta+d_\omega}}\nu^\epsilon(v)f(u-v)dv.\]
\end{definition}
By a simple change of variables and integration by part, one could derive that for any multi-index $J$, \[\nabla f^\epsilon=\nu^\epsilon*[D^Jf].\]
Here we quote some well-known results about this mollification from \cite[Theorem 7 of Appendix C.4]{Evans1998}.
\begin{lemma}\label{lem: molf}
\begin{enumerate}
  \item $f^\epsilon\in\mathcal C^{\infty}(\R^{d_\theta+d_\omega})$.
  \item $f^\epsilon\longrightarrow f$ almost everywhere as $\epsilon\longrightarrow0$. 
  \item If $f\in\mathcal C(\R^{d_\theta+d_\omega})$, then $f^\epsilon\longrightarrow f$ uniformly on compact subsets of $\R^{d_\theta+d_\omega}$.
  \item If $f\in\LL_{loc}^p(\R^{d_\theta+d_\omega})$ for some $1\leq p<\infty$, then $f^\epsilon\longrightarrow f$ in $\LL_{loc}^p(\R^{d_\theta+d_\omega})$.
\end{enumerate}
\end{lemma}
To give a convergence rate for the pointwise convergence in Lemma \ref{lem: molf}, we have the following proposition.
\begin{lemma}\label{lem: rho}
Assume $f\in W^{1,1}_{loc}(\R^{d_\theta+d_\omega})$ and there exist $k_1,k_2$ such that $|Df(u)|\leq k_1(1+\|u\|_2^{2k_2})$, then for any $u\in\R^{d_\theta+d_\omega}$, there exists $\rho:\R^+\to\R$ that $\lim_{\epsilon\to0}\rho(\epsilon)=0$ and $|f^\epsilon(u)-f(u)|\leq \rho(\epsilon)$.
\end{lemma}
\begin{proof}
\[\begin{aligned}
|f^\epsilon(u)-f(u)|&=\biggl|\int_{B_\epsilon(0)}\nu^\epsilon(v)[f(u-v)-f(u)]\biggl|dv\\
&=\biggl|\int_{B_\epsilon(0)}\nu^\epsilon(v)\int_0^1[Df(u-hv)^Tv]dhdv\biggl|\\
&\leq \epsilon\int_{B_\epsilon(0)}\nu^\epsilon(v)\int_0^1|Df(u-hv)|
dhdv\end{aligned}.\]
Since there exist $k_1,k_2$ such that $|Df(u)|\leq k_1(1+\|u\|_2^{2k_2})$,
\[\begin{aligned}
|f^\epsilon(u)-f(u)|&\leq\epsilon\int_{B_\epsilon(0)}\nu^\epsilon(v)\int_0^1\biggl[k_1(1+\|u-hv\|_2^{2k_2})\biggl]dhdv\\
&\leq\epsilon\int_{B_\epsilon(0)}\nu^\epsilon(v)\int_0^1\biggl[k_1(1+\|u\|_2^{2k_2}+h^{2k_2}\|v\|_2^{2k_2})\biggl]dhdv\\
&\leq\epsilon\int_{B_\epsilon(0)}\nu^\epsilon(v)\biggl[k_1(1+\|u\|_2^{2k_2})+\frac{k_1}{2k_2+1}\|v\|_2^{2k_2}\biggl]dv\\
&\leq\epsilon[k_1(1+\|u\|_2^{2k_2})]+\frac{k_1}{2k_2+1}\epsilon^{2k_2+1}.
\end{aligned}\]
Let $\rho(\epsilon)=\epsilon[k_1(1+\|u\|_2^{2k_2})]+\frac{k_1}{2k_2+1}\epsilon^{2k_2+1}$. Then $\rho(\epsilon)\longrightarrow0$ as $\epsilon\longrightarrow0$.\hfill$\square$
\end{proof}

It is also straightforward to see that mollification preserves Lipschitz conditions.

Consider the following SDE under component-wise mollification of coefficients,
\begin{equation}\label{eq: sde-mlf}\tag{SDE-MLF}
d\begin{pmatrix}\Theta_t^\epsilon\\\mathcal{W}_t^\epsilon\end{pmatrix}=[b_0^\epsilon(\Theta_t^\epsilon,\W_t^\epsilon)dt+\eta b_1^\epsilon(\Theta_t^\epsilon,\W_t^\epsilon)]+\sigma^\epsilon(\Theta_t^\epsilon,\W_t^\epsilon)dW_t.
\end{equation}
\begin{lemma}\label{lem: mlf-conv}
Assume $b_0$, $b_1$ and $\sigma$ are all Lipschitz. Then 
\[\E\biggl[\max_{t=1,\dots, N}\|\begin{pmatrix}\Theta_{t\eta}^\epsilon\\\mathcal{W}_{t\eta}^\epsilon\end{pmatrix}-\begin{pmatrix}\Theta_{t\eta}\\\mathcal{W}_{t\eta}\end{pmatrix}\|_2^2\biggl]\overset{\epsilon\to0}{\longrightarrow}0,\]
where $\begin{pmatrix}\Theta_{t\eta}^\epsilon\\\mathcal{W}_{t\eta}^\epsilon\end{pmatrix}$ and $\begin{pmatrix}\Theta_{t\eta}\\\mathcal{W}_{t\eta}\end{pmatrix}$ are given by \eqref{eq: sde-mlf} and \eqref{eq: gan-dyn}, respectively.
\end{lemma}
\begin{proof}
With Lemma \ref{lem: rho}, the conclusion follows from \cite[Theorem 9 in Section 2.5]{Krylov2008}.\hfill$\square$
\end{proof}

\subsubsection{Remaining proof}
Given the conditions of Theorem \ref{thm: gan-approx} and the fact that mollification preserves Lipschitz conditions, $b_0^\epsilon$, $b_1^\epsilon$ and $\sigma^\epsilon$ inherit regularity conditions from Theorem \ref{thm: gan-approx}. Therefore, the conclusion from Lemma \ref{lem: alt-sde-diff} holds. Lemmas \ref{lem: alt-diff}, \ref{lem: alt-sde-diff}, \ref{lem: moments} and \ref{lem: rho} verify the condition in \cite[Theorem 3]{Li2019}. Therefore, for any test function $f\in\mathcal C^3(\R^{d_\theta+d_\omega})$ such that for any multi-index $J$ with $|J|\leq 3$ there exist $k_1, k_2\in\mathbb N$ satisfying 
\[|\nabla^Jf(\theta,\omega)|\leq k_1\left(1+\left\|\begin{pmatrix}\theta\\\omega\end{pmatrix}\right\|_2^{2k_2}\right),\]
we have the following weak approximation,
\begin{equation}
\label{eq: alt-error}
\max_{t=1,\dots,N}\left|\E f(\theta_t,\omega_t)-\E f(\Theta_{t\eta}^\epsilon,\mathcal W_{t\eta}^\epsilon)\right|\leq C[\eta^2 + \rho(\epsilon)]
\end{equation}
for constant $C\geq0$, where $(\theta_t,\omega_t)$ and $(\Theta_{t\eta},\mathcal W_{t\eta})$ are given by \eqref{eq: dt-gan-update} and \eqref{eq: sde-mlf}, respectively, and $\rho$ is given as in Lemma \ref{lem: rho}.

Finally, taking $\epsilon$ to 0, Lemma \ref{lem: mlf-conv} and the explicit form of $\rho$ lead to the conclusion.

The proof of Theorem \ref{thm: sml-sde} can be executed in a similar fashion and thus omitted here.

\section{The long-run behavior of GANs training via invariant measure of SDE}\label{sec: conv}
In this section, we will study the long-run behavior of GANs training and discuss some of the implications of the technical assumptions as well as the limiting state.
\subsection{Long-run behavior of GANs training}
In addition to the evolution of parameters in GANs, 
the long-run behavior of GANs training can be estimated from these SDEs \eqref{eq: gan-dyn} and \eqref{eq: dyn-approx}. This limiting  behavior is characterized by their invariant measures. Recall the following definition of invariant measures in \citet{DaPrato2006}.
\begin{definition} \label{def: inv-msr}
A probability measure $\mu^*\in\mathcal P(\R^{d_\theta+d_\omega})$ is called an invariant measure for a stochastic process $\biggl\{\begin{pmatrix}\Theta_t\ \ \W_t\end{pmatrix}^T\biggl\}_{t\geq0}$ if for any measurable bounded function $f$ and $t\geq0$,
\[\int \E\left[f(\Theta_t,\W_t)|\Theta_0=\theta,\W_0=\omega\right]\mu^*(d\theta,d\omega)=\int f(\params)\mu^*(d\theta,d\omega).\]
\end{definition}
\begin{remark}
Intuitively, an invariant measure $\mu^*$ in the context of GANs training describes the joint probability distribution of the generator and discriminator parameters $(\Theta^*, \W^*)$ in equilibrium. For instance, if the training process converges to the unique minimax point $(\theta^*,\omega^*)$ for $\min_{\theta}\max_{\omega}\Phi(\params)$, the invariant measure is the Dirac mass at $(\theta^*,\omega^*)$. 

Moreover, the invariant measure $\mu^*$ and the marginal distribution of $\Theta^*$ characterize the generated distribution $Law(G_{\Theta^*}(Z))$, necessary for producing synthesized data and for evaluating the performance of the GANs model through metrics such as inception score and Fr\`echet inception distance. (See \citet{Salimans2016,Heusel2017} for more details on these metrics). 

Finally, as emphasized in Section \ref{sec: train} GANs are minimax game. From a game perspective,  the probability distribution of $\Theta^*$ given the discriminator parameter $\W^*$, denoted by the $Law(\Theta^*|\W^*)$, corresponds to the {\em mixed strategies} adopted by the generator; likewise,  the probability distribution of $\W^*$ given the generator parameter $\Theta^*$, denoted by $Law(\W^*|\Theta^*)$, characterizes the {\em mixed strategies} adopted by the discriminator.
\end{remark}


Recall that the SDE approximation \eqref{eq: gan-dyn} for GANs training process is given by
\[d\begin{pmatrix}\Theta_t\\\mathcal{W}_t\end{pmatrix}=b(\Theta_t,\W_t)dt+\sigma(\Theta_t,\W_t)dW_t,\]
where the drift coefficient is given by $b(\params)=b_0(\params)+\eta b_1(\params)$ with
\begin{align*}
    b_0(\params)&=\begin{pmatrix}-\gt(\params)\\\go(\params)\end{pmatrix},\\
    b_1(\params)&=\frac{1}{2}\begin{pmatrix}\nabla_\theta\gt(\params)&-\nabla_\omega\gt(\params)\\-\nabla_\theta\go(\params)&-\nabla_\omega\go(\params)\end{pmatrix}\begin{pmatrix}-\gt(\params)\\\go(\params)\end{pmatrix}\\
    &=-\frac{1}{2}\nabla b_0(\params)b_0(\params)-\begin{pmatrix}\nabla_\omega\gt(\params)\go(\params)\\0\end{pmatrix},
\end{align*}
and the diffusion coefficient is given by
\[\sigma(\params)=\sqrt{2\beta^{-1}}\begin{pmatrix}\sigt(\Theta_t,\W_t)^{\frac{1}{2}}&0\\
    0&\sigo(\Theta_t,\W_t)^{\frac{1}{2}}\end{pmatrix}.\]
Note that \eqref{eq: gan-dyn} depends on the first and second order derivatives of the training loss with respect to the generator and the discriminator parameters.
\begin{theorem}
\label{thm: inv-msr}
Assume the following conditions hold: 
\begin{enumerate}[label=\ref{thm: inv-msr}.\alph*]
    \item both $b$ and $\sigma$ are bounded and smooth and have bounded derivatives of any order;
    \item there exist some positive real numbers $r$ and $M_0$ such that for any $\begin{pmatrix}\theta\ \ \omega\end{pmatrix}^T\in\R^{d_\theta+d_\omega}$,
    \[\begin{pmatrix}\theta\ \ \omega\end{pmatrix} b(\params)\leq -r\left\|\begin{pmatrix}\theta\\\omega\end{pmatrix}\right\|_2,\text{ if }\left\|\begin{pmatrix}\theta\\\omega\end{pmatrix}\right\|_2\geq M_0;\]
    \item $\A$ is uniformly elliptic, i.e., there exists $l>0$ such that for any $\begin{pmatrix}\theta\\\omega\end{pmatrix},\begin{pmatrix}\theta'\\\omega'\end{pmatrix}\in\R^{d_\theta+d_\omega}$,
    \[\begin{pmatrix}\theta'\ \ \omega'\end{pmatrix}^T\sigma(\params)\sigma(\params)^T\begin{pmatrix}\theta'\\\omega'\end{pmatrix}\geq l\left\|\begin{pmatrix}\theta'\\\omega'\end{pmatrix}\right\|_2^2,\] 
\end{enumerate}
then \eqref{eq: gan-dyn}  admits a unique invariant measure $\mu^*$ with an exponential convergence rate. 

Similar results hold for the invariant measure of  \eqref{eq: dyn-approx} with $b$ replaced by $b_0$. 
\end{theorem}
\paragraph{Proof of Theorem \ref{thm: inv-msr}.}
In order to prove Theorem \ref{thm: inv-msr}, we will construct an appropriate Lyapunov function to characterize the long-term behavior for the SDE \eqref{eq: gan-dyn}; the associated Lyapunov condition leads to the existence of an invariant measure for the dynamics of the parameters. We highlight this very technique since it can be used in the analysis of broader classes of dynamical systems, for both stochastic and deterministic cases; see for instance \citet{Laborde2019}. Consider the following function $V:[0,\infty)\times\R^{d_\theta+d_\omega}\to\R$,
\begin{equation}\label{eq: lyapunov}\tag{Lyapunov}
V(t,u)=\exp\{\delta t+\epsilon\|u\|_2\},\quad \forall u\in\R^{d_\theta+d_\omega},
\end{equation}
where the parameters $\delta, \epsilon>0$ will be determined later. Note that $V$ is a smooth function, and 
\begin{equation}\label{eq: ext-1}\lim_{\|u\|_2\to\infty}\inf_{t\geq0}V(t,u)=+\infty,\end{equation}
for any fixed $\delta,\epsilon>0$. Under \eqref{eq: gan-dyn}, applying It\^o's formula to $V$ gives
\begin{align*}
&dV(t,\Theta_t,\W_t)=V(t,\Theta_t,\W_t)\left[\epsilon\frac{\begin{pmatrix}\Theta_t\ \ \W_t\end{pmatrix} b(\Theta_t,\W_t)}{\left\|\begin{pmatrix}\Theta_t\ \ \W_t\end{pmatrix}^T\right\|_2}+\delta +\frac{1}{2}Tr\left(\sigma(\Theta_t,\W_t)\sigma(\Theta_t,\W_t)^T\right.\right.\\
&\left.\left.{}\times\left[\frac{\epsilon\left\|\begin{pmatrix}\Theta_t\ \ \W_t\end{pmatrix}^T\right\|_2^2I+(\epsilon^2\left\|\begin{pmatrix}\Theta_t\ \ \W_t\end{pmatrix}^T\right\|_2-\epsilon)\begin{pmatrix}\Theta_t\ \ \W_t\end{pmatrix}^T\begin{pmatrix}\Theta_t\ \ \W_t\end{pmatrix}}{\left\|\begin{pmatrix}\Theta_t\ \ \W_t\end{pmatrix}^T\right\|_2^3}\right]\right)\right]dt\\
&+\epsilon V(t,\Theta_t,\W_t)\frac{\begin{pmatrix}\Theta_t\ \ \W_t\end{pmatrix}\sigma(\Theta_t,\W_t)}{\left\|\begin{pmatrix}\Theta_t\ \ \W_t\end{pmatrix}^T\right\|_2}dW_t.
\end{align*}
Define the Lyapunov operator
\[\LL V(t,u)=V(t,u)\biggl[\epsilon\frac{u^Tb(u)}{\|u\|_2}+\delta+\frac{1}{2}Tr\biggl(\sigma(u)\sigma(u)^T\frac{\epsilon\|u\|_2^2I+(\epsilon^2\|u\|_2-\epsilon)uu^T}{\|u\|_2^3}\biggl)\biggl].\]
Given the boundedness of $\sigma$, i.e. there exists $K>0$ such that $\|\sigma\|_F\leq K$, and dissipative property given by condition 2, i.e. there exists $l,M_0>0$ such that for any $u\in\R^{d_\theta+d_\omega}$ with $\|u\|_2>M_0$,
\[u^Tb(u)\leq -l\|u\|_2,\]
we have that 
\[\LL V(t,u)\leq V(t,u)\left[\delta-l\epsilon+\frac{1}{2}\biggl(\epsilon\frac{\|\sigma\|_F^2}{\|u\|_2}+\epsilon^2\|\sigma\|_F^2\biggl)\right]\leq V(t,u)\left[\delta+\frac{K^2\epsilon^2}{2}-\left(l-\frac{K^2}{2\|u\|_2}\right)\epsilon\right].\]
Now take $M>\max\left\{\frac{K^2}{2l},M_0\right\}$, $0<\epsilon<\frac{2l}{K^2}-\frac{1}{M}$ and $\delta=-\frac{1}{2}\left[\frac{K^2\epsilon^2}{2}+\left(\frac{K^2}{2M}-l\right)\epsilon\right]>0$, then for any $\|u\|_2>M$,
\[\LL V(t,u)\leq -\delta V(t,u).\]
Therefore,
\begin{equation}\label{eq: ext-2}\lim_{\|u\|_2\to\infty}\inf_{t\geq0}\LL V(t,u)=-\infty.\end{equation}
Following \cite[Theorem 3.7]{Khasminskii2011}, \eqref{eq: ext-1} and \eqref{eq: ext-2} ensure the existence of a invariant measure $\mu^*$ for \eqref{eq: gan-dyn}. By the uniform elliptic condition \ref{thm: inv-msr}.c, uniqueness follows from \cite[Theorem 2.3]{Hong2019}. The exponential convergence rate follows from \cite[Main result]{Veretennikov1988}.
\subsection{Discussions}
\paragraph{Implications of the technical assumptions on GANs training.}
The assumptions \ref{thm: gan-approx}.a--\ref{thm: gan-approx}.c, \ref{thm: sml-sde}.a-\ref{thm: sml-sde}.b and \ref{thm: inv-msr}.a for the regularity conditions of the drift, the volatility, and the derivatives of loss function $\Phi$, are more than  mathematical convenience. They are essential constraints on the growth of the loss function with respect to the model parameters, necessary for avoiding the explosive gradient encountered in the training of GANs.

Moreover, these conditions put restrictions on the gradients of the objective functions with respect to the parameters. By the chain rule, it requires both careful choices of network structures as well as particular forms of the loss function $\Phi$.

In terms of  proper neural network architectures, let us take an example of a network with one hidden layer. Let $f:\X\subset\R^{d_x}\to\R$ with
\[f(x;W^h,w^o)=\sigma_o\left(w^o\cdot \boldsymbol{\sigma}_h(W^hx)\right)=\sigma_o\left(\sum_{i=1}^h w_i^o\sigma_h\left(\sum_{j=1}^{d_x}W_{i,j}^hx_j\right)\right).\]
Here $h$ is the width of the hidden layer, $W^h\in\R^{h\times d_x}$ and $w^o\in\R^h$ are the weight matrix and vector for the hidden and output layers respectively, and $\sigma_h:\R\to\R$ and $\sigma_o:\R\to\R$ are the activation functions for the hidden and output layers. Then taking the partial derivatives with respects to the weights yields
\[\begin{aligned}
&\partial_{W^h_{i,j}}f(x;W^h,w^o)=\sigma_o'\left(\sum_{i=1}^h w_i^o\sigma_h\left(\sum_{j=1}^{d_x}W_{i,j}^hx_j\right)\right)\cdot w_i^o\cdot \sigma_h'\left(\sum_{j=1}^{d_x}W_{i,j}^hx_j\right)\cdot x_j,\\
&\partial_{w^o_i}f(x;W^h,w^o)=\sigma_o'\left(\sum_{i=1}^h w_i^o\sigma_h\left(\sum_{j=1}^{d_x}W_{i,j}^hx_j\right)\right)\cdot \sigma_h\left(\sum_{j=1}^{d_x}W_{i,j}^hx_j\right),
\end{aligned}\]
from which we can see that the regularity conditions, especially the growth of the loss function with respect to the model parameters (i.e., assumptions 1.a-1.c and 2.a-2.b) rely on regularity and boundedness of the activation functions, the width and depth of the network, as well as the magnitudes of parameters and data. Therefore, assumptions 1.a-1.c, 2.a-2.b and 3.a explain mathematically some well-known practices in GANs training such as introducing various forms of gradient penalties; see for instance \citet{Gulrajani2017, Thanh-Tung2019}. See also \cite{Schafer2019} for a combination of competition and gradient penalty to stabilize GANs training. It is worth noting that apart from affecting the stability of GANs training, the regularity of the network can also affect the sample complexity of GANs and this phenomenon has been studied in \citet{Luise2020} for a class of GANs with optimal transport-based loss functions.
    
In terms of choices of loss functions,  the objective function of the vanilla GANs  in \citet{Goodfellow2014} is given by
\[l(\params)=\E_{X\sim\pbm_r}[\log D_\omega(X)]+\E_{Z\sim\pbm_z}[\log(1-D_\omega(G_\theta(Z)))].\]Take the partial derivatives with respect to $\theta$ and $\omega$, we see
\[\begin{aligned}
&\nabla_\theta l(\params)=-E_{Z\sim\pbm_z}\left[\frac{1}{1-D_\omega(G_\theta(Z))}\mathbf{J}^G_\theta(Z)\nabla_x D_\omega(G_\theta(Z))\right],\\
&\nabla_\omega l(\params)=\E_{X\sim\pbm_r}\left[\frac{1} {D_\omega(X)}\nabla_\omega D_\omega(X)\right]-\E_{Z\sim\pbm_z}\left[\frac{1}{1-D_\omega(G_\theta(Z))}\nabla_\omega D_\omega(G_\theta(Z))\right],
\end{aligned}\]
where $\mathbf J^G_\theta$ denotes the Jacobian matrix of $G_\theta$ with respect to $\theta$ and $\nabla_x$ denotes the gradient operator over a (parametrized) function with respect to its input variable. \citet{Arjovsky2017a} analyzed the difficulties of stabilizing GANs training under above loss function due to the lack of its regularity conditions, and proposes a possible remedy by an alternative  Wasserstein distance which are smoother functions.

\paragraph{Verifiability of  assumptions in Theorems \ref{thm: gan-approx}, \ref{thm: sml-sde} and \ref{thm: inv-msr}.} These assumptions can be summarized into three categories specified below. There are many choices of GANs structures for a wide range of applications where these assumptions can be verified easily. Specifically,
\begin{enumerate}
    \item {\bf On smoothness and boundedness of drift and volatility.} Take the example of  WGANs for image processing. Given that sample data in image processing problems are supported on compact domain, then assumptions 1.a-1.c, 2.a-2.b and 3.a  are easily satisfied with proper prior distribution and activation function: first, the prior distribution $\mathbb P_z$ such as the uniform distribution is naturally compactly supported; next, take $D_\omega=\tanh{(\omega\cdot x)}$,  $G_\theta(z)=\tanh{(\theta\cdot z)}$, and the objective function
    \[\Phi(\theta, \omega)=\frac{\sum_{i=1}^N\sum_{j=1}^M D_\omega(x_j)-D_\omega(G_\theta(z_i))}{N\cdot M}.\]
    Then the assumptions 1.a-1.c, 2.a-2.b and 3.a are guaranteed by boundedness of the data $\{(z_i,z_j)\}_{1\leq i\leq N,1\leq j\leq M}$ and property of \[\psi(y)=\tanh{y}=\frac{e^y-e^{-y}}{e^y+e^{-y}}=1-\frac{2}{e^{2y}+1}\in(-1,1).\] More precisely, the first and second order derivatives of $\psi$ are \[\psi'(y)=\frac{4}{(e^y+e^{-y})^2}\in(0,1],\quad \psi''(y)=-8\frac{e^y-e^{-y}}{(e^y+e^{-y})^3}=-2\psi(y)\psi'(y)\in(-2,2).\]
    Any higher order derivatives can be written as functions of $\psi(\cdot)$ and $\psi'(\cdot)$ and therefore bounded.
    \item {\bf On the dissipative property.} The dissipative property specified by 3.b essentially prevents the evolution of the parameters from being driven to infinity. The weights clipping technique in WGANs, for instance, is consistent with this assumption.
    \item {\bf On the elliptic condition}. The elliptic property of volatility term specified in 3.c is trivially satisfied given its expression in \eqref{eq: volatility}. 
\end{enumerate}

\subsection{Dynamics of training loss and FDR}
One can further analyze the dynamics of the training loss based on the SDE approximation; and derive a fluctuation-dissipation relation (FDR) for the GANs training.  

To see this,  let $\mu=\{\mu_t\}_{t\geq0}$ be the flow of probability measures for $\left\{\begin{pmatrix}\Theta_t\\\mathcal W_t\end{pmatrix}\right\}_{t\geq0}$ given by \eqref{eq: gan-dyn}. Then
applying It\^o's formula to the smooth function $\Phi$ \citep[see][sec.~4.18]{Rogers2000} gives the following dynamics of training loss,
\begin{equation}\label{eq: ito}\Phi(\Theta_t,\W_t)=\Phi(\Theta_s,\W_s)+\int_s^t \A\Phi(\Theta_r,\W_r)dr +\int_s^t\sigma(\Theta_r,\W_r)\nabla\Phi(\Theta_r,\W_r)dW_r;
\end{equation}
where 
\begin{equation}
    \label{eq: generator}
    \A f(\params) =  b(\params)^T \nabla f(\params)+ \frac{1}{2}Tr\left(\sigma(\params)\sigma(\params)^T\nabla^2f(\params)\right),
\end{equation} 
is the infinitesimal generator for \eqref{eq: gan-dyn} on given any test function $f:\R^{d_\theta+d_\omega}\rightarrow\R$. 

The existence of the unique invariant measure $\mu^*$ for  \eqref{eq: gan-dyn} implies the convergence of $\begin{pmatrix}\Theta_t\\\mathcal W_t\end{pmatrix}$ in \eqref{eq: gan-dyn}  to some $\begin{pmatrix}\Theta^*\\\mathcal W^*\end{pmatrix}\sim\mu^*$, as $t\to\infty$. By the Definition \ref{def: inv-msr} of invariant measure and \eqref{eq: ito}, we have
\[\E_{\mu^*}[\A\Phi(\Theta^*,\W^*)]=0.\]
Apply the operator \eqref{eq: generator} over the loss function $\phi$ yields
\[\begin{aligned}
    \A\Phi(\params)=&b_0(\params)^T\nabla\Phi(\params)+\eta b_1(\params)^T\nabla\Phi(\params)+\frac{1}{2}Tr(\sigma(\params)\sigma(\params)^T\nabla^2\Phi(\params))\\
    =&-\|\nabla_\theta\Phi(\params)\|_2^2+\|\nabla_\omega\Phi(\params)\|_2^2\\
    &-\frac{\eta}{2}\biggl[\nabla_\theta\Phi(\params)^T\nabla_\theta^2\Phi(\params)\nabla_\theta\Phi(\params)+\nabla_\omega\Phi(\params)^T\nabla_\omega^2\Phi(\params)\nabla_\omega\Phi(\params)\biggl]\\
    &+\beta^{-1}Tr\biggl(\sigt(\params)\nabla_\theta^2\Phi(\params)+\sigo(\params)\nabla_\omega^2\Phi(\params)\biggl).
\end{aligned}\]
Based on the evolution of loss function \eqref{eq: ito}, convergence to the invariant measure $\mu^*$ leads to the following 
FRD for GANs training. 
\begin{theorem}
\label{thm: fdr}
Assume the existence of an invariant measure $\mu^*$ for \eqref{eq: gan-dyn}, then 
\begin{equation}
    \label{eq: fdr}\tag{FDR1}
    \begin{aligned}
      &\E_{\mu^*}\biggl[\|\nabla_\theta\Phi(\Theta^*,\W^*)\|_2^2-\|\nabla_\omega\Phi(\Theta^*,\W^*)\|_2^2\biggl]=\beta^{-1}\E_{\mu^*}\biggl[Tr\biggl(\sigt(\Theta^*,\W^*)\nabla_\theta^2\Phi(\Theta^*,\W^*)\\
      &\hspace{15pt}+\sigo(\Theta^*,\W^*)\nabla_\omega^2\Phi(\Theta^*,\W^*)\biggl)\biggl]-\frac{\eta}{2}\E_{\mu^*}\biggl[\nabla_\theta\Phi(\Theta^*,\W^*)^T\nabla_\theta^2\Phi(\Theta^*,\W^*)\nabla_\theta\Phi(\Theta^*,\W^*)\\
      &\hspace{15pt}+\nabla_\omega\Phi(\Theta^*,\W^*)^T\nabla_\omega^2\Phi(\Theta^*,\W^*)\nabla_\omega\Phi(\Theta^*,\W^*)\biggl].
    \end{aligned}
\end{equation}
The corresponding FDR for the simultaneous update case of  \eqref{eq: dyn-approx} is
\begin{equation*}
    \label{eq: fdr-sml}
    \begin{aligned}
    &\E_{\mu^*}\biggl[\|\nabla_\theta\Phi(\Theta^*,\W^*)\|_2^2-\|\nabla_\omega\Phi(\Theta^*,\W^*)\|_2^2\biggl]=\\
   &\hspace{20pt}\beta^{-1}
   \E_{\mu^*}\biggl[Tr\biggl(\sigt(\Theta^*,\W^*)\nabla_\theta^2\Phi(\Theta^*,\W^*)+\sigo(\Theta^*,\W^*)\nabla_\omega^2\Phi(\Theta^*,\W^*)\biggl)\biggl].
   \end{aligned}
\end{equation*}
\end{theorem}

\begin{remark}
This FDR relation in GANs connects the microscopic fluctuation from the noise of SGA with the macroscopic dissipation phenomena related to the loss function. In particular, the quantity $Tr(\sigt\nabla^2_\theta\Phi+\sigo\nabla^2_\omega\Phi)$  links the covariance matrices $\sigt$ and $\sigo$ from SGAs with the loss landscape of $\Phi$, and reveals the trade-off of the loss landscape between the generator and the discriminator. 

Note that this FDR relation for GANs training is analogous to  that for stochastic gradient descent  algorithm on a pure minimization problem in \citet{Yaida2019} and \citet{Liu2019b}.
\end{remark}

Further analysis of the invariant measure can lead to a different type of FDR that will be practically useful for learning rate scheduling. 
Indeed,  applying It\^o's formula to the squared norm of the parameters
$\biggl\|\begin{pmatrix}\Theta_t\\\W_t\end{pmatrix}\biggl\|_2^2$ shows the
following dynamics \[d\biggl\|\begin{pmatrix}\Theta_t\\\W_t\end{pmatrix}\biggl\|_2^2=2\begin{pmatrix}\Theta_t\\\W_t\end{pmatrix}^Td\begin{pmatrix}\Theta_t\\\W_t\end{pmatrix}+Tr\biggl(\sigma(\Theta_t,\W_t)\sigma(\Theta_t,\W_t)^T\biggl)dt.\]
\begin{theorem}
\label{thm: fdr2}
Assume the existence of an invariant measure $\mu^*$ for \eqref{eq: dyn-approx}, then
\begin{equation}
    \label{eq: fdr2}\tag{FDR2}
   \E_{\mu^*}\biggl[\Theta^{*,T}\nabla_\theta\Phi(\Theta^*,\W^*)-\W^{*,T}\nabla_\omega\Phi(\Theta^*,\W^*)\biggl]=\beta^{-1}\E_{\mu^*}\biggl[Tr(\sigt(\Theta^*,\W^*)+\sigo(\Theta^*,\W^*))\biggl]
\end{equation}
\end{theorem}
Given  the infinitesimal generator for \eqref{eq: gan-dyn}, Theorems \ref{thm: fdr} and \ref{thm: fdr2} follow from direct computations. 
\begin{remark}[Scheduling of learning rate]
Notice that the quantities in \eqref{eq: fdr2}, including the parameters $(\params)$ and first-order derivatives of the  loss function $\gt$, $\go$, $\gtij$ and $\goij$, are computationally inexpensive. Therefore,\eqref{eq: fdr2}
enables customized scheduling of learning rate, 
instead of  predetermined scheduling ones such as Adam or RMSprop optimizer. 

For instance, recall that $\gtb$ and $\gob$ are
respectively unbiased estimators for $\gt$ and $\go$, and 
\begin{align*}
&\hat\Sigma_\theta(\params)=\frac{\sum_{k=1}^B[\gtbk(\params)-\gtb(\params)][\gtbk(\params)-\gtb(\params)]^T}{B-1},\\
&\hat\Sigma_\omega(\params)=\frac{\sum_{k=1}^B[\gobk(\params)-\gob(\params)][\gobk(\params)-\gob(\params)]^T}{B-1}
\end{align*}
are respectively unbiased estimators of $\sigt(\params)$ and $\sigo(\params)$. Now in order  to improve GANs training  with the simultaneous update, one can introduce two tunable parameters $\epsilon>0$ and $\delta>0$ to have the following scheduling:
\begin{center}
if $\left|\frac{\Theta^T\gtb(\Theta_t,\W_t)-\W_t^T\gob(\Theta_t,\W_t)}{\beta^{-1}Tr(\hat\Sigma_\theta(\Theta_t,\W_t)+\hat\Sigma_\omega(\Theta_t,\W_t))}-1\right|<\epsilon$, then update $\eta$ by $(1-\delta)\eta$.
\end{center}
\end{remark}

Proofs of Theorems \ref{thm: fdr} and \ref{thm: fdr2} are deferred to the Appendix.

\bibliography{references}

\end{document}